\documentclass[10pt, conference, compsocconf]{IEEEtran}

\usepackage{multirow} 
\usepackage{subcaption}
\usepackage{url}
\usepackage{multirow}
\usepackage{hyperref}
\usepackage{amsfonts} 
\usepackage{amsmath} 
\usepackage{comment}

\ifCLASSINFOpdf
  \usepackage[pdftex]{graphicx}
\else
\fi
\def\IEEEbibitemsep{0pt plus .5pt}

\begin{document}
%
\title{Feature Density Estimation for Out-of-Distribution Detection via Normalizing Flows}



\author{\IEEEauthorblockN{Evan D. Cook, Marc-Antoine Lavoie, Steven L. Waslander}
\IEEEauthorblockA{Institute for Aerospace Studies\\
University of Toronto\\
Toronto, ON, Canada\\
\{evan.cook, marc-antoine.lavoie, steven.waslander\}@robotics.utias.utoronto.ca}}


%


\maketitle

\begin{abstract}
Out-of-distribution (OOD) detection is a critical task for safe deployment of learning systems in the open world setting. In this work, we investigate the use of feature density estimation via normalizing flows for OOD detection and present a fully unsupervised approach which requires no exposure to OOD data, avoiding researcher bias in OOD sample selection. This is a post-hoc method which can be applied to any pretrained model, and involves training a lightweight auxiliary normalizing flow model to perform the out-of-distribution detection via density thresholding. Experiments on OOD detection in image classification show strong results for far-OOD data detection with only a single epoch of flow training, including 98.2\% AUROC for ImageNet-1k vs. Textures, which exceeds the state of the art by 7.8\%. We additionally explore the connection between the feature space distribution of the pretrained model and the performance of our method. Finally, we provide insights into training pitfalls that have plagued normalizing flows for use in OOD detection.

\end{abstract}

\begin{IEEEkeywords}
normalizing flows; out of distribution detection;

\end{IEEEkeywords}

%
\IEEEpeerreviewmaketitle

\section{Introduction}

Machine learning has rapidly advanced in recent years, with state of the art models performing impressive tasks in a wide range of technical domains. However the standard workflow in machine learning is significantly less flexible than learning observed in animals in nature. While biological neural systems continually learn in uncontrolled environments, artificial neural networks are instead trained with a closed-world assumption \cite{OodSurvey} on a fixed corpus of training data, validated against a set of reserved data drawn from the same data distribution, and then deployed to perform roughly the same task. When deployed, these models can be exposed to inputs that are dissimilar to the in-distribution (ID) data they were trained and validated on, potentially leading to unpredictable behavior when encountering this out-of-distribution (OOD) data.

Addressing how artificial neural networks can be used in open world situations where they may be exposed to out-of-distribution data remains a challenge. Out-of-distribution detection is the task of identifying when a sample is not drawn from the training data distribution. This is especially important in safety critical applications such as autonomous vehicles; the statistical assurances on model performance provided by the validation dataset are no longer applicable.

In this work, we revisit using feature density estimation (FDE) via normalizing flows for out-of-distribution detection in image classification. Prior works assert that normalizing flows are not effective for OOD detection when performing density estimation in pixel space \cite{DoGenerativeModelsKnowWhatTheyDontKnow}, and density estimation in the feature space of pretrained models has been discussed but not thoroughly investigated \cite{FlowsFailOOD}. We demonstrate that by performing density estimation in the feature space of a pretrained image classification backbone model, normalizing the feature representations, and under-training the normalizing flow we are able to achieve competitive results on both small and large datasets. The proposed method has the advantages of being fully unsupervised and requires no exposure to OOD training data, avoiding researcher bias from a specific definition of the OOD data. Finally, this is a post-hoc method that can be applied to any pretrained classification model, and it requires training a lightweight normalizing flow model for only a single epoch to perform the feature-space density estimation for out-of-distribution detection, making it a broadly applicable technique.

\section{Related Work}
    
\subsection{Out-of-Distribution Detection}

OOD detection deals with identifying semantically distinct samples (from unseen classes) to avoid erroneously classifying them as one of the classes in the training distribution. Out-of-distribution detection performance is evaluated by attempting to discriminate between a validation dataset versus an out-of-distribution dataset. The most widely used metric is the area under the receiver operating characteristic (AUROC)~\cite{AUROC}, a threshold-free classification performance metric useful for comparing unbalanced datasets. 

OOD detection is a rich field with many existing approaches. These can be divided into classification-based, distance-based, generative-based, and density-based methods \cite{OodSurvey}. Classification-based approaches define a classification output that identifies ID and OOD inputs at inference time, with common baseline methods including the max-softmax probability (MSP) \cite{hendrycks2018baseline}, ODIN \cite{ODIN}, the energy score \cite{EnergyOOD}, and post-hoc methods that modify the feature vector activations such as ASH \cite{ASH} and ReAct \cite{ReAct}. MSP is a simple baseline method which thresholds on the maximum class probability. Energy score is a more modern development with stronger performance while remaining simple to implement, calculating a metric inspired by thermodynamics (the free energy) from the classification logits. ReAct is used in conjunction with the energy score, but clips off the top 10\% of feature vector activations prior to evaluation, resulting in state of the art performance on large scale datasets. Distance-based methods label OOD samples as those sufficiently far from ID training samples (in feature space), and include Euclidean and Mahalanobis distance \cite{Mahalanobis}. Generative-based approaches employ generative models to reconstruct inputs, and assess samples with poor reconstruction or low likelihood under the generative model as OOD \cite{UnifiedOodSurvey}. Examples of this approach include VAEs with modified priors \cite{tilted_vae}, hierarchical VAEs \cite{hierarchical_vae}, and diffusion models \cite{Diffusion_Graham_2023_CVPR, liu2023unsupervised}. These methods require training a generative network to model the data distribution, which can be computationally demanding.

Finally, for density-based approaches a density estimation model is built from the training data such that the ID data lies within high density regions, and OOD data encountered at inference time occupies low density regions. A threshold on the density can be added to transform a density estimator into an out-of-distribution detector, classifying low probability data as out-of-distribution. In this approach the density estimator is used as a proxy for model epistemic uncertainty \cite{Uncertainty}. Density estimation methods can be performed in the input data space or a transformed representation space, and include kernel methods, radial basis functions, and normalizing flows \cite{Uncertainty, OodSurvey}. It was observed in \cite{FlowsFailOOD} that using a normalizing flow to perform density estimation in the feature space improves performance over density estimation in the pixel space, but their analysis is extremely limited and their results do not reach other state-of-the-art methods. 

\subsection{Normalizing Flows}

\begin{figure*}[t]
\centering
\includegraphics[width=0.9\textwidth]{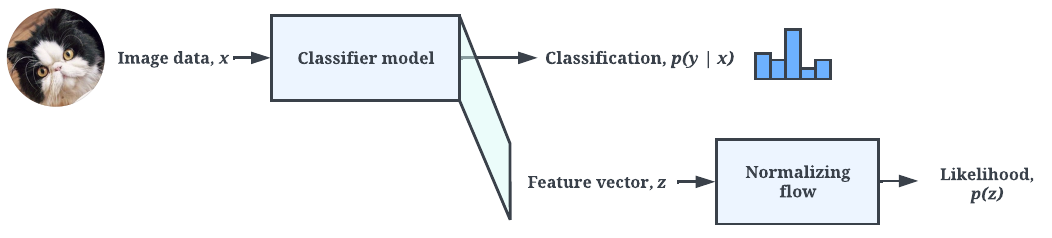}
\caption{Feature density estimation for out-of-distribution detection.}
\label{method}
\end{figure*}

\begin{figure}[t]
    \centering
    \subcaptionbox{}{
        \includegraphics[width=0.45\linewidth]{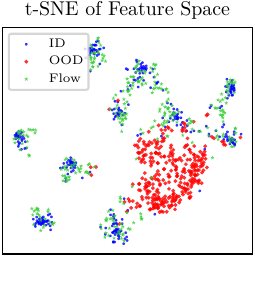}
    }
    \subcaptionbox{}{
        \includegraphics[width=0.45\linewidth]{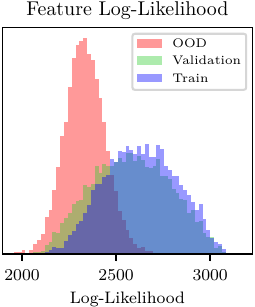}
    }
    \caption{(a) A t-SNE visualization of the feature space  of a ResNet18 backbone model comparing ID data (CIFAR-10), OOD data (SVHN), and flow generated feature vectors. Clearly visible are 10 clusters corresponding to the 10 classes of CIFAR-10, and the coincidence between the ID data and the flow generated data points. (b) A histogram of the log-likelihood of ID features vs. OOD features under the normalizing flow model. Better separability of these distributions leads to higher AUROC for OOD detection.}
    \label{tsne}
\end{figure}

Normalizing flows are a class of generative neural networks that are dimensionality preserving and fully invertible. They are trained to learn a diffeomorphism to map between two probability distributions, typically a data distribution and a known base probability distribution (such as the normal distribution). Normalizing flows have the dual function of being an exact density estimator (by measuring the probability of a datapoint mapped to the base distribution), and a generative model (by sampling from the base distribution, and then mapping into the data space). Mathematically, normalizing flows can be written as implementing a change of variables:
\[p(z) = q(f_{\theta}(z)) \left| \det \left( \frac{\partial f_{\theta}(z)}{\partial z^{T}} \right) \right| \]
%
where $p(z)$ is the data distribution, $q(z)$ is the known base distribution, and $f_{\theta}(z)$ is the mapping function between these two distributions, implemented as an invertible normalizing flow neural network parameterized by $\theta$.

For a more thorough and formal review of the mathematics of normalizing flows, we refer readers to \cite{FlowReview}. Implementing normalizing flows is often challenging, as the model must be entirely invertible and should have a Jacobian that can be efficiently calculated. However, they have shown impressive performance in many tasks, including generating realistic images of faces \cite{Glow} and high quality density estimation on image data \cite{FlowPlusPlus}.

\subsection{Normalizing Flows for Out-of-Distribution Detection}

Normalizing flows have been applied to the task of OOD detection in several prior works with mixed success, but have historically performed very poorly for OOD detection in the image classification domain. When performing density estimation on pixel data in images, previous authors recommend against the use of normalizing flows, finding that they learn spurious pixel-level correlations and capture low-level statistics rather than high-level semantics \cite{DoGenerativeModelsKnowWhatTheyDontKnow, FlowsFailOOD, UnderstandingGenerativeFailures}. 

In \cite{Gudovskiy_2022_WACV}, \cite{Rudolph_2021_WACV}, and \cite{Rudolph_2022_WACV} normalizing flows are applied to image segmentation anomaly detection by performing density estimation of multiscale feature map embeddings instead of pixel space. Results are promising, but limited to small scale datasets, and they use hand-tailored network architectures that do not generalize to other domains.

Flows have also been used for anomaly detection in video data. A Glow normalizing flow \cite{Glow} is used by \cite{VideoAnomalyDetection} to perform density estimation of the feature vectors produced by two autoencoders, one capturing spatial information and one capturing temporal information. This work highlights the importance of performing density estimation in the feature space and demonstrates competitive performance in this domain, but has a complex autoencoder architecture with a reconstruction loss term, limiting its potential applications. In \cite{waymo_flow_mining} a normalizing flow is applied to the task of quantifying sample rareness, illustrating the value of feature space density estimation with normalizing flows for the downstream task of data mining and dataset balancing.

In \cite{HybridModelsForOpenSet} strong OOD detection performance is demonstrated using normalizing flows in image classification, but this is not a post-hoc method, as it requires jointly training  the classifier backbone and normalizing flow together with additional hyperparameters. This approach is limited by the necessity to jointly learn the feature space, and results are only evaluated on small datasets. In \cite{FlowsFailOOD}, the concept of performing density estimation in the feature space of a pretrained classifier is briefly discussed, but their analysis is very limited and their OOD performance is not compelling. Our work carries the investigation of feature density estimation via normalizing flows much further, demonstrating that normalizing flows can achieve state of the art out-of-distribution detection in image classification using a simple, post-hoc method with no complex architecture changes or modifications to the backbone.

\section{Method}

\subsection{Feature Density Estimation}

In this work we leverage a pretrained neural network backbone to provide a compressed, reduced representation of our input data that is rich in semantic information for the downstream task of image classification. We use the penultimate layer's activations as feature vectors for density estimation. These feature vectors contain all of the necessary information for the backbone model to perform the output classification task, and are typically transformed to the final output logits using a linear projection head.

We perform density estimation on the feature representations using established normalizing flow architectures \cite{Glow, realNVP, ResidualFlow}, learning an invertible mapping between the feature space and a normal probability distribution. Our normalizing flows are trained on the penultimate layer activations of a frozen pretrained image classifier, with the optimization criterion of minimizing the log-likelihood of the transformed features. As an unsupervised method, the class labels of the original image data are unused. Once trained, the normalizing flow is a computationally efficient density estimator for the feature space of the pretrained backbone model. Out-of-distribution detection is achieved by applying a simple probability threshold to the density estimates for new samples, classifying low density samples as OOD. See Figures \ref{method} and \ref{tsne} for a block diagram and visualization of our method.

\section{Experimental Setup}
\label{section:experiments}

We evaluate the utility of normalizing flows for out-of-distribution detection on a range of image classification tasks, using a variety of backbone networks and in-distribution datasets. 

\textbf{Datasets}: We use CIFAR-10 \cite{cifar} and ImageNet-1k \cite{imagenet} as our in-distribution datasets. CIFAR-10 models are evaluated against random Gaussian noise, SVHN \cite{SVHN}, Places365 \cite{Places}, CelebA \cite{CelebA}, and CIFAR-100 \cite{cifar} as out-of-distribution datasets. ImageNet-1k models are evaluated against Textures \cite{Textures} and reduced versions of iNaturalist \cite{iNaturalist}, SUN \cite{SUN}, and Places \cite{Places} datasets. The latter three are filtered to ensure these datasets contain no common classes with ImageNet-1k, as done by \cite{ReAct}. Datasets can be considered as either \textit{near-OOD} or \textit{far-OOD} \cite{UnifiedOodSurvey} depending on how semantically distinct they are from the ID dataset; in this work we consider Gaussian noise and SVHN to be far-OOD from CIFAR-10, and Textures to be far-OOD from ImageNet-1k.

\textbf{Evaluation}: Out-of-distribution detection performance is evaluated using AUROC, calculated between the in-distribution validation dataset and out-of-distribution dataset. AUROC is a threshold-free metric, and an AUROC of 50\% indicates no separability between the distributions, while an AUROC of 100\% indicates perfect separability between the distributions. We evaluate our method against MSP \cite{hendrycks2018baseline}, ODIN \cite{ODIN}, energy score \cite{EnergyOOD}, and ReAct \cite{ReAct}. 

\textbf{Normalizing Flow Models}: We use a 10 block Glow \cite{Glow} flow for all experiments. For flow models trained on CIFAR-10, each block is composed of two linear layers with dimension [512, 2048, 512]. For flow models trained on ImageNet-1k, each block is composed of two linear layers which do not alter the dimensionality of the feature space (2048 for ResNet50 and 768 for Swin-T). Flow models are trained using the Adam optimizer \cite{Adam} for only a single epoch with a learning rate of 1e-4 for CIFAR-10 and 1e-5 for ImageNet-1k.

\textbf{Backbone Model}: For CIFAR-10, we train a ResNet18 classifier backbone using supervised learning to a validation accuracy of 92.1\%. For ImageNet-1k, we use two PyTorch pretrained models as classifier backbones: ResNet50 and Swin-T, with top-1 validation accuracies of 76.1\% and 81.5\% respectively \cite{pytorch_pretrained}. Backbone weights are frozen for all experiments.

\section{Results}

We first present our main results for CIFAR-10, summarized Table \ref{table:cifar_table}. Our method exceeds other approaches on SVHN and Gaussian noise (far-OOD), and is competitive with alternatives on more challenging datasets (Places365, CelebA, CIFAR-100).
Further, we present results for the larger scale ImageNet-1k dataset in Table \ref{table:imagenet_table}. With a ResNet backbone, our method is able to achieve 98.2\% AUROC on Textures \cite{Textures}, obtaining 7.8\% better performance than the next best method ReAct \cite{ReAct}. With a transformer backbone, we again outperform ReAct by 7.5\% on Textures. Our method consistently outperforms competing methods at detecting the more visually distinct far-OOD samples (CFIAR-10 vs. SVHN, and ImageNet vs. Textures).

\begin{table*}[htbp]
    \centering
    \caption{Out-of-distribution detection performance results, with CIFAR-10 as in-distribution.}
    \label{table:cifar_table}
    \begin{tabular}{p{2cm}p{2cm}ccccc}
        \hline
        \textbf{\multirow{2}{*}{Backbone}} & \textbf{\multirow{2}{*}{Method}} & \multicolumn{5}{c}{\textbf{Out-of-Distribution Dataset (AUROC $\uparrow$)}} \\
        \cline{3-7}
        && \textbf{Gaussian} & \textbf{SVHN} & \textbf{Places365} & \textbf{CelebA} & \textbf{CIFAR-100} \\
        \hline
        \multirow{4}{*}{\shortstack[l]{ResNet18\\ + CIFAR-10}} & MSP & 90.86 & 88.62 & 88.78 & 90.62 & 85.86 \\
                 & Energy Score & 87.33 & 87.12 & 92.77 & 94.62 & \textbf{88.35} \\
                 & ODIN & 88.95 & 87.47 & \textbf{95.13} & \textbf{95.16} & 87.43 \\
                 & FDE (ours) & \textbf{99.41} & \textbf{96.06} & 92.61 & 93.23 & 85.93 \\
        \hline
    \end{tabular}
\end{table*}

\begin{table*}[htbp]
    \centering
    \caption{Out-of-distribution detection performance results, with ImageNet-1k as in-distribution.}
    \label{table:imagenet_table}
    \begin{tabular}{p{4cm}p{2.5cm}cccc}
        \hline
        \textbf{\multirow{2}{*}{Backbone}} & \textbf{\multirow{2}{*}{Method}} & \multicolumn{4}{c}{\textbf{Out-of-Distribution Dataset (AUROC $\uparrow$)}} \\
        \cline{3-6}
        && \textbf{Textures} & \textbf{iNaturalist} & \textbf{SUN} & \textbf{Places} \\
        \hline
        \multirow{5}{*}{ResNet50 + ImageNet-1k} & MSP & 80.49 & 88.39 & 81.64 & 80.53 \\
                 & Energy Score & 86.80 & 90.62 & 86.57 & 83.96 \\
                 & ReAct & 90.44 & \textbf{96.43} & \textbf{94.42} & \textbf{91.93} \\

                 & ODIN & 87.72 & 91.41 & 86.82 & 84.36 \\
                 & FDE (ours) & \textbf{98.20} & 81.75 & 78.61 & 72.46
 \\
        \hline
        \multirow{5}{*}{Swin-T + ImagetNet-1k} & MSP & 81.72 & 89.86 & 81.26 & \textbf{80.33} \\
               & Energy Score & 78.05 & 85.13 & 73.34 & 70.60 \\
               & ReAct & 85.48 & 90.01 & 81.50 & 79.28 \\
               & ODIN & 67.70 & 71.13 & 57.84 & 54.57 \\
               & FDE (ours) & \textbf{92.94} & \textbf{92.15} & \textbf{81.96} & 77.02 \\
        \hline
    \end{tabular}
\end{table*}

\section{Discussion}

To practically implement normalizing flows for OOD detection, we discuss several key considerations.

\subsection{Flow Regularization}
\label{section:flow_regularization}

\begin{figure}[h]
    \centering
    \subcaptionbox{Epoch 0, AUROC=96.1\%}{
        \includegraphics[width=0.46\linewidth]{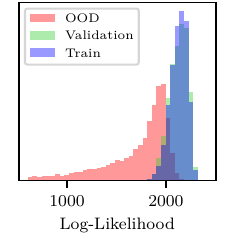}
    }
    \subcaptionbox{Epoch 999, AUROC=84.4\%}{
        \includegraphics[width=0.46\linewidth]{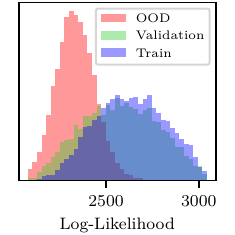}
    }
    \caption{Feature likelihood histograms for the same flow model at 0 epochs and 999 epochs. With further training, the likelihood increases for all distributions, but the training and validation distributions begin to separate due to overfitting, while the separability of the ID/OOD distributions degrades.}
    \label{figure:epoch_histogram_comparison}
\end{figure}

During training, the goal is to optimize a flow model that fits the training distribution, generalizes to the validation set, and still separates OOD data. It is critically important to manage overfitting: the separability of validation and OOD data is directly impacted by the normalizing flow’s generalization gap between the training and validation data distributions (see Figure \ref{figure:epoch_histogram_comparison} for a visualization of training, validation, and OOD likelihood distributions).

A variety of standard regularizations techniques can be used to avoid overfitting on the training data. Especially important is data augmentation. We train our flows on feature vectors obtained with dataset augmentation identical to those used to train the backbone model.

\begin{figure}[h]
    \centering
    \includegraphics[width=0.46\linewidth]{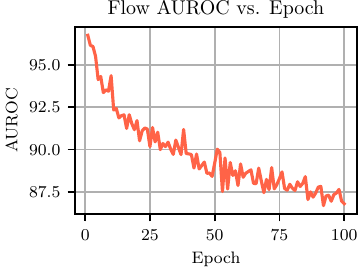}
    \includegraphics[width=0.46\linewidth]{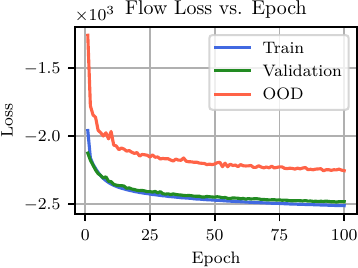}
    \caption{Normalizing flow AUROC (CIFAR-10 vs. SVHN) and loss during training. Loss of ID \textit{and} OOD test datasets decreases during training. AUROC peaks early then declines.}
    \label{figure:training_curve}
\end{figure}

Under-training was found to be critically important. Our experiments demonstrated a surprising trend: training a normalizing flow model to minimize the validation loss may actually be detrimental to OOD detection performance. Counterintuitively, loss on a test OOD dataset \textit{decreases} during training (OOD data becomes more likely as the flow model fits to ID data), and AUROC for this test OOD dataset peaks early, then drops with additional epochs as the flow model fits to the training data (see Figure \ref{figure:training_curve}). The optimal number of epochs to train a flow model for depends on the OOD dataset, flow architecture, and backbone model, and is far lower than when the validation loss begins to rise (classic overfitting). This is thus distinct from early stopping, and represents a novel and beneficial form of under-training. In our work, we report all results using normalizing flows trained to only a single epoch to ensure consistent evaluations across models and datasets.

\subsection{Normalizing Feature Vectors}

\begin{figure}[t]
    \centering
        \includegraphics[width=\linewidth]{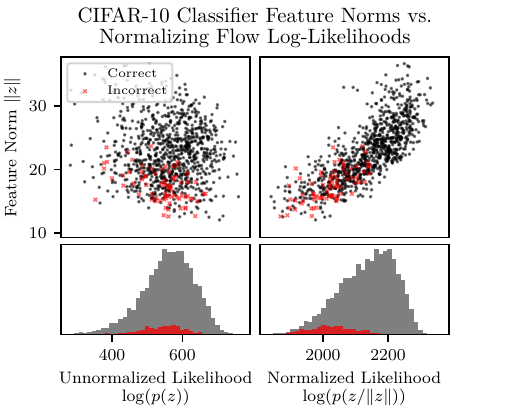}
        \caption{Visualization of feature vector norms vs. log-likelihood for a flow model trained with normalized (left) and unnormalized (right) feature vectors. For a flow model trained on unnormalized features, there is no correlation between feature norm, classification accuracy, and flow likelihood. For a flow model trained on normalized features, a correlation is observed between feature norm, classification accuracy, and flow likelihood.}
    \label{feature_norms}
\end{figure}

We find that normalizing the feature vectors strongly improves OOD detection performance on far-OOD datasets (CIFAR-10 vs. SVHN and ImageNet-1k vs. Textures). We can write the final linear head of the classification model as:
\[y = W^Tz = \|z\| (W^T \hat{z})\]
where $W$ is the weight matrix (ignoring the bias term), $z$ is our feature vector, $\|z\|$ is the Euclidean norm of $z$, and $\hat{z}$ is the normalized (unit-length) feature vector. We interpret the product $W^T \hat{z}$ as the semantic agreement between $\hat{z}$ and each logit's class, while $\|z\|$ relates to the network’s overall confidence in the output. Larger feature norms, $\|z\|$, correlate with larger logits and higher classification probabilities. Training a flow density estimator on unnormalized feature vectors confounds the interpretation of the log-likelihood of the features. Features with very large norms may be modelled as having a low likelihood (see Figure \ref{feature_norms}), which is counterproductive. The OOD detection task is concerned with OOD data that are outside the expected semantic content, not outside the expected data likelihood (a perfect image of a cat may be semantically ID, but could be considered OOD due to an unusually high feature vector norm).  We resolve this by training the normalizing flow model on normalized feature vectors: density estimation is performed on the semantic content of the feature space disentangled from the classifier's confidence. As shown in Figure \ref{feature_norms}, training our flow on normalized features yields a strong correlation between feature norms and feature log-likelihood: this can be interpreted as a desirable correlation between classifier confidence and likelihood of the semantic content of the feature. Further, training on normalized features introduces a correlation between feature likelihood and the probability of correct classification (despite this being an unsupervised method with no access to classification labels); no such correlation exists when training on unnormalized features. Experiments show that OOD detection performance is substantially improved on far-OOD data when performing density estimation on normalized features (Tables \ref{table:cifar_scaling_table}, \ref{table:imagenet_scaling_table}).

\begin{table*}[htbp]
    \centering
    \caption{Flow AUROC results, CIFAR-10: normalized vs. unnormalized features}
    \label{table:cifar_scaling_table}
    \begin{tabular}{p{4cm}p{2.5cm}cc}
        \hline
        \textbf{\multirow{2}{*}{Backbone Model}} & \textbf{\multirow{2}{*}{Feature Scaling}} & \multicolumn{2}{c}{\textbf{Out-of-Distribution Dataset (AUROC $\uparrow$)}} \\
        \cline{3-4}
        && \textbf{SVHN} & \textbf{CIFAR-100}  \\
        \hline
        \multirow{2}{*}{ResNet18 + CIFAR-10} & Unnormalized & 85.63 & 57.68 \\
                 & Normalized & \textbf{96.48} & \textbf{85.93} \\
        \hline
    \end{tabular}
\end{table*}

\begin{table*}[htbp]
    \centering
    \caption{Flow AUROC results, ImageNet-1k: normalized vs. unnormalized features}
    \label{table:imagenet_scaling_table}
    \begin{tabular}{p{4cm}p{2.5cm}cccc}
        \hline
        \textbf{\multirow{2}{*}{Backbone Model}} & \textbf{\multirow{2}{*}{Feature Scaling}} & \multicolumn{4}{c}{\textbf{Out-of-Distribution Dataset (AUROC $\uparrow$)}} \\
        \cline{3-6}
        && \textbf{Textures} & \textbf{iNaturalist} & \textbf{SUN} & \textbf{Places} \\
        \hline
        \multirow{2}{*}{ResNet50 + ImageNet-1k} & Unnormalized & 86.95 & 49.78 & 60.21 & 58.89 \\
                 & Normalized & \textbf{98.20} & \textbf{81.75} & \textbf{78.61 }& \textbf{72.46} \\
        \hline
        \multirow{2}{*}{Swin-T + ImageNet-1k} & Unnormalized & 87.87 & 89.27 & \textbf{82.27} & \textbf{79.56} \\
               & Normalized & \textbf{92.94} & \textbf{92.15} & 81.96 & 77.02 \\
        \hline
    \end{tabular}
\end{table*}

\subsection{Flow Architecture}

Normalizing flow architecture is an active area of research, with different flow designs having their own pros and cons. RealNVP \cite{realNVP} is fast and simple flow architecture but performs poorly compared to more modern methods. Glow \cite{Glow} demonstrates good performance as a generative model but is not state of the art for density estimation, and Residual Flows \cite{ResidualFlow} are excellent density estimators but are slower to train than alternatives.

\begin{table*}[th]
    \centering
    \caption{Normalizing flow OOD detection vs. architecture comparison. AUROC results are generally comparable, and flow models with superior density estimation do not equate to improved OOD detection performance.}
    \label{table:flow_architecture}
    \begin{tabular}{p{3.5cm}p{3cm}cc}
        \hline
        \textbf{\multirow{2}{*}{Backbone Model}} & \textbf{\multirow{2}{*}{Flow Architecture}} & \multicolumn{2}{c}{\textbf{Out-of-Distribution Dataset (AUROC $\uparrow$)}} \\
        \cline{3-4}
        && \textbf{SVHN} & \textbf{CIFAR-100}  \\
        \hline
        \multirow{3}{*}{ResNet18 + CIFAR-10} 
                 & RealNVP & 92.13 & 86.33 \\
                 & Glow & 96.99 & 85.87 \\
                 & Residual Flow & 96.89 & 80.32 \\
        \hline
    \end{tabular}
\end{table*}

Surprisingly, our experiments show that the performance of OOD detection is relatively insensitive to flow architecture. We believe this is due to the fact that discrimination between two distributions (the validation dataset and OOD dataset) is the key task, rather than high quality modeling of the training distribution. Maximum OOD detection performance is often seen after only a few training epochs of the flow model, far before the training distribution is adequately modelled.

As discussed in Section \ref{section:flow_regularization}, extensively training the flow model to maximize the likelihood of the ID data is unimportant for OOD detection. Instead, the difference in likelihood between the ID and OOD distributions is more important. As such, more sophisticated flow models which advance the state of the art in density estimation and offer improved likelihood of ID data are not necessarily advantageous for OOD detection (see Table \ref{table:flow_architecture}). In our experiments Glow \cite{Glow} was used, as it performed well while being faster than more complex methods, such as Residual flows \cite{ResidualFlow} and FFJORD \cite{FFJORD}, and was stable to train.

\subsection{Backbone Feature Distribution}

OOD detection performance is independent of classification accuracy, and is strongly affected by the distribution of feature representations produced by the backbone model. OOD detection performance may vary wildly between different classifiers of similar accuracy, and understanding which innate properties of neural networks (including model architectures, training hyperparameters, pretraining data distributions, and pretraining loss) improve the OOD detection task is understudied compared to the primary task of improving classification accuracy.

To investigate the factors that influence a backbone model's OOD detection performance, we examine the feature space distribution using two metrics: uniformity and tolerance \cite{contrastive_2021_CVPR}.
\[\text{Uniformity}(z;t) = \log \mathbb{E}_{x,y\sim p_{\text{data}}}\left[e^{-t\|z(x)-z(y)\|_2^2}\right]\]
\[\text{Tolerance}(z) = \mathbb{E}_{x,y\sim p_{\text{data}}}\left[\left(z(x)^T z(y)\right) \cdot \mathbf{1}_{\{l(x)=l(y)\}}\right]\]
Here, $l(x)$ is the supervised label of datapoint $x$. Uniformity uses a Gaussian similarity kernel to measure the spread of features over the feature space hypersphere. We use a weight parameter of $t=2$, and report negative uniformity consistent with \cite{contrastive_2021_CVPR}. Higher uniformity means more of the hypersphere is occupied by features produced by the model. Tolerance measures the cosine similarity of intra-class feature representations: tolerance is higher when class representations form tight clusters in feature space, and is lower when class representations are more diffuse in feature space. To investigate the connection between feature space distribution and the ability to detect OOD data using feature density estimation via a normalizing flow, we apply our method to 80 pretrained PyTorch ImageNet-1k classification models \cite{pytorch_pretrained}. We  evaluate the uniformity and tolerance of each classifier's feature space for ID data, train a normalizing flow on this feature space, and evaluate the performance of our OOD detection method.

\begin{figure}[t]
    \centering
        \includegraphics[width=\linewidth]{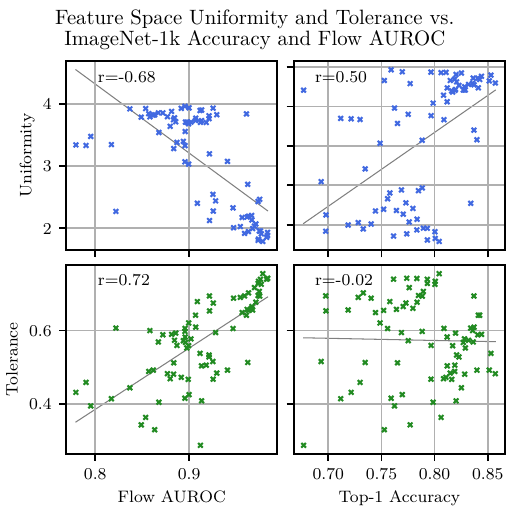}
    \caption{Evaluating 80 pretrained PyTorch ImageNet-1k classifiers for feature space uniformity and tolerance, validation accuracy, and flow AUROC (ImageNet-1k vs. Textures). The Pearson correlation coefficient is annotated.}
    \label{uniformity_tolerance}
\end{figure}

Visualizing the performance of our method (evaluated as ImageNet-1k vs. Textures AUROC) versus the uniformity and tolerance of 80 unique feature spaces (Figure \ref{uniformity_tolerance}) shows two strong correlations: AUROC is positively correlated with tolerance (tight class clustering, Pearson correlation coefficient $r = 0.72$), and negatively correlated with uniformity (volume of feature space occupied, $r = -0.68$). Our method is thus best applied to models with compact ID class representations occupying a lower volume of feature space. It is easier to fit a normalizing flow density model to these distributions, and they have an increased likelihood of OOD samples falling in the low density regions. 

Additionally, we see that uniformity is correlated with classifier top-1 validation accuracy ($r = 0.50$), while no strong correlation exists between tolerance and top-1 validation accuracy ($r = -0.02$, Figure \ref{uniformity_tolerance}). A tradeoff is thus apparent for uniformity: high uniformity correlates with improved classification accuracy, but low uniformity improves OOD detection using feature density estimation. An ideal model may balance this by exhibiting high uniformity and high tolerance, encouraging both validation accuracy and OOD detection. Future work will investigate the role of training regularizations, model architectures, and pretraining loss and task in creating feature spaces amenable to both strong OOD detection and classification accuracy.

\section{Conclusion}

For machine learning systems to be safely deployed in the open world, it is essential that out-of-distribution data can be accurately identified to safeguard against unintended model behavior. We investigate a method for out-of-distribution detection by performing density estimation in the feature space of pretrained image classification models using normalizing flows. In contrast with prior work in this space, our experiments show that feature density estimation via normalizing flows can achieve strong OOD detection performance on a variety of common benchmarks on large scale datasets. Our method outperforms all existing methods for detecting far-OOD data, as demonstrated by the results on CIFAR-10 vs. SVHN, and ImageNet-1k vs. Textures.

Performing density estimation on normalized feature vectors and under-training the normalizing flow are shown to be particularly important, and we observe the surprising behavior that OOD detection performance peaks very early in flow training. We further show that OOD detection performance is not dependent on the flow model's ability to perform high quality density estimation, but is strongly dependent on the distribution of feature representations of the backbone model. Specifically, evaluations of 80 pretrained ImageNet-1k classifiers show that performance of our method is strongly correlated with the tolerance of the classifier's feature space. Using the discussed techniques, we demonstrate that normalizing flows are effective tools for OOD detection, blazing a trail towards the safe deployment of machine learning and robotic systems in challenging open world environments.




%

\bibliographystyle{IEEEtran}
\bibliography{main}

\end{document}